\documentclass[review]{elsarticle}

\usepackage{lineno,hyperref}
\modulolinenumbers[5]

\usepackage{graphics,graphicx,epsfig,algorithm,algpseudocode}
\usepackage{latexsym,amsmath,amsfonts,amssymb,color}
\usepackage{graphicx}
\usepackage{url,textcomp}
\usepackage{subcaption}

\newcommand\tabcaption{\def\@captype{table}\caption}
\newcommand{\be}{\begin{eqnarray*}}
\newcommand{\ee}{\end{eqnarray*}}
\newcommand{\ibe}{\begin{eqnarray}}
\newcommand{\iee}{\end{eqnarray}}

\journal{Computers \& Mathematics with Applications}

\begin{document}

\begin{frontmatter}

\title{Learning finite difference methods for reaction-diffusion type equations with FCNN}

\author[a]{Yongho Kim}
\ead{ykim@mpi-magdeburg.mpg.de}

\author[b]{Yongho Choi\corref{mycorrespondingauthor}}
\cortext[mycorrespondingauthor]{Corresponding author}
\ead{yongho\_choi@daegu.ac.kr}
\ead[url]{https://sites.google.com/view/yh-choi}

\address[a]{Faculty of Mathematics, Otto-von-Guericke-Universität Magdeburg, Universitätsplatz 2, 39106, Magdeburg, Germany}
\address[b]{Department of Computer \& Information Engineering, Daegu University, Gyeongsan-si, Gyeongsangbuk-do 38453, Republic of Korea}

\begin{abstract}
In recent years, Physics-informed neural networks (PINNs) have been widely used to solve partial differential equations alongside numerical methods because PINNs can be trained without observations and deal with continuous-time problems directly. In contrast, optimizing the parameters of such models is difficult, and individual training sessions must be performed to predict the evolutions of each different initial condition. To alleviate the first problem, observed data can be injected directly into the loss function part. To solve the second problem, a network architecture can be built as a framework to learn a finite difference method. In view of the two motivations, we propose Five-point stencil CNNs (FCNNs) containing a five-point stencil kernel and a trainable approximation function for reaction-diffusion type equations including the heat, Fisher’s, Allen--Cahn, and other reaction-diffusion equations with trigonometric function terms. We show that FCNNs can learn finite difference schemes using few data and achieve the low relative errors of diverse reaction-diffusion evolutions with unseen initial conditions. Furthermore, we demonstrate that FCNNs can still be trained well even with using noisy data.

\end{abstract}

\begin{keyword}
convolutional neural networks \sep reaction-diffusion type equations \sep five-point stencil CNN
\MSC[2010] 65M06 \sep 68T01
\end{keyword}

\end{frontmatter}


\section{Introduction}
To express diverse natural phenomena such as sound, heat, electrostatics, elasticity, thermodynamics, fluid dynamics, and quantum mechanics mathematically, various partial differential equations (PDEs) have been derived and numerical methods can be applied to solve these PDEs. Representative numerical methods for solving PDEs include the finite difference method, finite element method, finite volume method, spectral method, and so on. In this study, we focus on the finite difference method (FDM) which divides a given domain into finite grids and finds an approximate solution using derivatives with finite differences \cite{PZ2012,JDSY2017}. This method uses each points and its neighbors to predict the corresponding point at the next time step. Likewise, in convolutional neural networks (CNNs) \cite{cnn}, convolution operators extract each pixel of outputs using the corresponding pixel and its neighbor pixels of an input. Also, the convolution operators are generally immutable. Hence, well-structured CNNs have the potential to solve partial differential equations numerically \cite{cnnallencahn}. Among various PDEs representing natural phenomena, we consider reaction-diffusion type equations. The reaction-diffusion model has been applied and used in various fields such as biology \cite{NB1986,PBH2016,DYYMDJJJ2017}, chemistry \cite{BAG2009,ISBBDL2012,GHRR2013}, image segmentation \cite{HBB1995,SYHR2006,ZYMQS2020},  image inpainting \cite{MBet2000,YDJSJ2015,JJS2016}, medicine \cite{EQ2015,HYJ2015,MCYSA2021}. In this paper, we focus on second order reaction-diffusion type equations, including the heat, Fisher's, Allen--Cahn (AC) equation, and reaction-diffusion equations with trigonometric function terms.

In recent years, neural networks have been widely applied to solve PDEs. As a popular framework, Physics-informed neural networks (PINNs) \cite{pinns1} based on multi-layer perceptrons (MLPs) approximate PDE solutions by the optimization of a loss function including given laws of physics. The primary advantage of PINNs is that PDE solutions can be inferred without any iterative process such as a recurrence equation with respect to time. Furthermore, it is used for diverse applications such as Hidden Fluid Mechanics (HFMs) \cite{hfm} that extract hidden variables of a given equation using a PINN and observations. However, training models depending on intricate PDEs and their coefficients is classically difficult. Also, the inference of a PINN depends on a given initial condition, hence an individual training session is required whenever the initial condition changes. To improve the training ability, PINNs have been combined with numerical methods as well as other neural networks such as CNNs have been selected. M. Raissi et al. \cite{pinns1} added Runge-Kutta methods to a PINN model to solve the AC equation. Aditi et al. \cite{pinns2} proposed transfer learning and curriculum regularization which start training PINNs on a specific safe domain and then transfer their learning  to a target domain. Hao Ma et al. \cite{unetpde}  proposed a U-shaped CNN called U-net \cite{unet} and showed that the usage of target data in the loss function part significantly improves the model optimization. Elie Bretin et al. \cite{meancur} used convolutional neural networks derived from a semi-implicit approach to learn phase field mean curvature flows of the AC equation. We propose Five-point stencil CNNs (FCNNs) containing a five-point stencil kernel and a trainable approximation function to obtain the numerical solutions of second order reaction-diffusion type equations. Our contributions are as follows: 
\begin{enumerate}
    \item {We propose a five-point stencil convolution operator to solve reaction-diffusion type equations.}
    \item {We show that finite difference methods can be reconstructed by FCNNs with two consecutive snapshots and that FCNNs achieve low relative errors for diverse evolutions.}
    \item{We demonstrate that the robustness of FCNNs using five reaction-diffusion type equations and noisy data.}
\end{enumerate}

The remainder of this paper is organized as follows. In Section \ref{sec:methods}, we present how to create training data using explicit FDMs and explain the concept of FCNNs, training process, and numerical solutions. In Section \ref{sec:results1} and \ref{sec:results2}, we measure the relative errors between FCNN and FDM solutions as well as we show the robustness of FCNNs using the diverse initial conditions and noisy data. In Section \ref{sec:results3}, we compare FCNN to PINN. Finally, summarizes our results and concludes the work in Section \ref{sec:conclusions}.

\section{Methods and numerical solutions} \label{sec:methods}
The FDM divides a given domain into finite grids and find an approximate solution using derivatives with finite differences \cite{PZ2012,JDSY2017}. To create training data for each equation, we apply an explicit FDM to create training data with a random initial condition. A computational domain is defined using a uniform grid of size $h=1/N_x=1/N_y$ and $\Omega_h = \{ (x_i, y_j)=(a+(i-1.5)h, c+(j-1.5)h) \}$ for $1 \leq i \leq N_x +2$ and $1 \leq j \leq N_y +2$. Here, $N_x$ and $N_y$ are mesh sizes on the computational domain $(a,b) \times (c,d)$. Let $\phi_{ij}^n$ be approximations of $\phi(x_i, y_j, n\Delta t)$ and $\Delta t$ be temporal step size. The boundary condition is a zero Neumann boundary condition. The Laplacian of a function $\phi$ is calculated using a five-point stencil method, the Laplacian $\bigtriangleup \phi$ can be approximated as follows:
\begin{equation}\label{Lpde}
\bigtriangleup_h \phi \approx\frac{\phi(x+h,y)+\phi(x-h,y)-4\phi(x,y)+\phi(x,y+h)+\phi(x,y-h)}{h^2}.
\end{equation}

In this way, the first and second derivatives of $\phi$ at each point $(x_i,y_j)$ (e.g., $\phi_x$, $\phi_y$, $\phi_{xx}$ and $\phi_{yy}$) can be approximated within the 3 $\times$ 3 local area centered $(x_i,y_j)$. This concept can be equivalent to 3 $\times$ 3 convolution kernels. The 3 $\times$ 3 kernels $K$ following properties: 

1. $k_1\oplus k_2\in K$ for any $k_1, k_2\in K$ (element-wise summation).

2. $k_1\odot k_2\in K$ for any $k_1, k_2\in K$ (element-wise multiplication).

3. $k^{-1}\in K$ for any $k\in K$ (element-wise division).

4. $a k\in K$ for any $k\in K$ and any real numbers $a$.

Therefore, second-order PDEs can be solved numerically by the properties of 3 $\times$ 3 kernels \cite{cnnallencahn}.

To solve second order reaction-diffusion type equations 
\begin{equation}
\phi_t=\alpha\bigtriangleup\phi+\beta f(\phi),
\end{equation}
where $\alpha$ is a diffusion coefficient, $\beta$ is a reaction coefficient, and $f$ is a smooth function to present reaction effect. We propose FCNN as a recurrence relation:
\begin{equation}\label{rec2}
\phi^{n+1} =\phi^n +\Delta t \alpha\bigtriangleup_h\phi^n+\Delta t \beta f(\phi^n).
\end{equation}

As a CNN, $F(\phi^n)$ containing a 5-point stencil filter and a pad satisfying given boundary conditions solves $\Delta t \alpha\bigtriangleup\phi^n$. To approximate $\phi^n +\Delta t\beta f(\phi^n)$ terms, we define a trainable polynomial function $\epsilon(\phi^n)$ as follows:
\begin{equation}
\epsilon(\phi^n)=a_0+\sum_{k=1}^{N} a_i(\phi^n-b)^k, \label{poly}
\end{equation}
with model parameters $a_i$ for any $i \in \{0,1,\cdots,N\}$ and a real value $b$. Let $M(\phi^n) = F(\phi^n) + \epsilon(\phi^n)$ be an FCNN. Then, the inference is performed as follows:
\begin{equation}\label{receq3}
\phi^{n+1} = M_{\theta}(\phi^n),
\end{equation}
where $\theta$ is a set of model parameters. 

 \begin{figure}[htbp!]
\centering
\includegraphics[width=0.5\columnwidth]{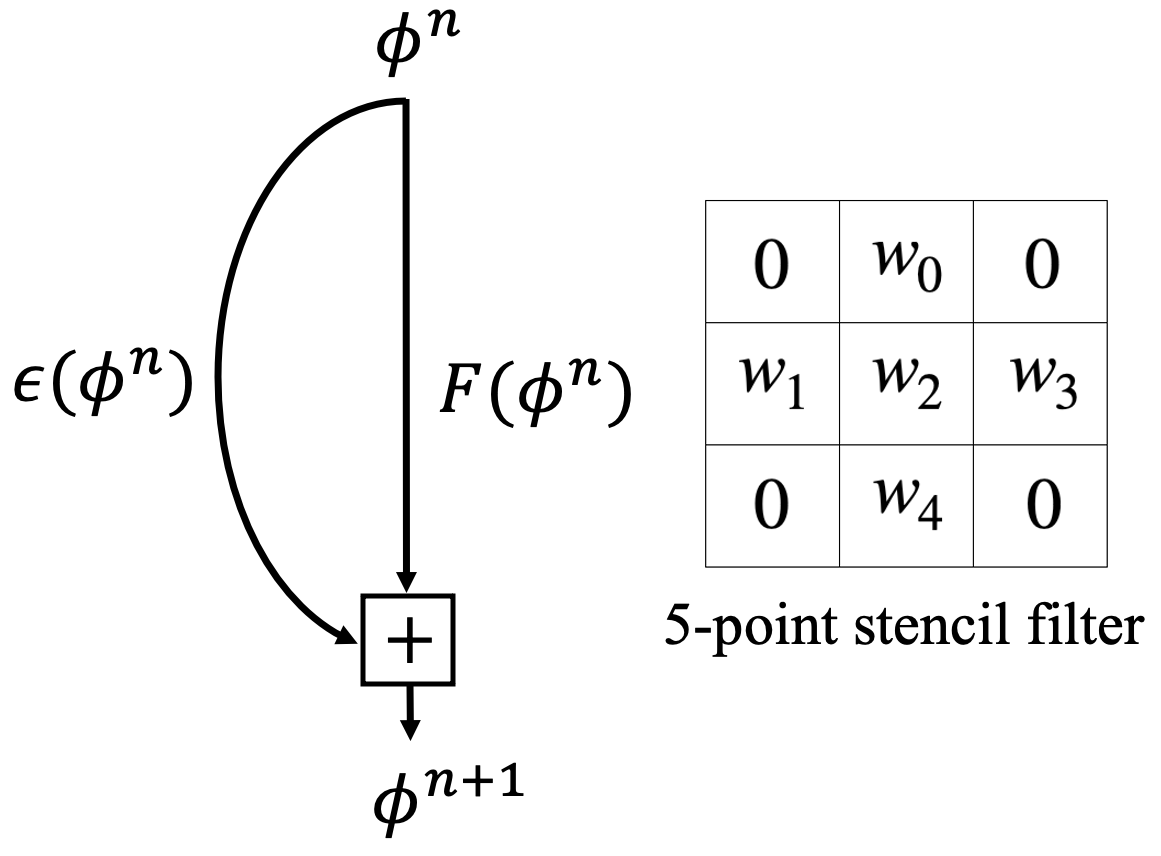}
\caption{Computational graph of FCNN for second order reaction-diffusion type equations}%
\label{modelg}
\end{figure}

Figure \ref{modelg} shows the computational graph of our explicit model FCNN containing model parameters $w_i$ for any $i \in \{0,1,2,3,4\}$ in a filter.  Furthermore, $F$ represents the diffusion term on the uniform grid of $x$ and $y$ axises, so we set up $w_1=w_3$ and $w_0=w_4$  to reduce training time. When the five-point stencil filter is used and $\epsilon$ is a $p$-th order polynomial function, the number of model parameters is only $p+4$. Thus, the set-up enables to learn physical patterns from few data. In Algorithm \ref{fcnn},  an initial image $\phi^{0}(=u^{0})$ and the prediction $\phi^{1}$ at the next time $\Delta t$ are used with training data $u^0$ and $u^1$ to train a model $M_{\theta}$. The objective function $L(\phi^1, u^1)$ is the mean square error function without physics-informed loss as follows: 
\begin{equation}\label{receq3}
L(\phi^1,u^1)=\frac{1}{N}\sum_{i=1}^{N}( \phi_i^1-u_i^1)^2,
\end{equation}
  where $N$, $\phi^{1}$ and $u^{1}$ are the number of pixels in an output image, a prediction and its target respectively.

\begin{algorithm}[htbp!]
\caption{Training Procedure}
\label{fcnn}
\begin{algorithmic}

    \State Set an initial value $\phi^0=u^0$, a small constant $\delta>0$
    \State Initialize $M_{\theta}(\phi^0) = F(\phi^0) + \epsilon(\phi^0)$ with model parameters $\theta$
    
    \While{$\ell > \delta$}  
        \State $\phi^{1} \leftarrow M_{\theta}(\phi^0)$  
        \State Compute loss $\ell=L(\phi^1, u^1)$
        \State Update $\theta$
    \EndWhile  

\end{algorithmic}
\end{algorithm}

\subsection{Reaction-diffusion type equations}
To demonstrate the robustness of FCNNs, we consider reaction-diffusion type equations including the heat, Fisher's, AC equation, reaction-diffusion equations with trigonometric function terms. The reaction and diffusion coefficients used in each formula are shown in Table \ref{coeftab}.
\begin{table}[htbp]
 \begin{center}
  \caption{Diffusion ($\alpha$) and reaction ($\beta$) coefficients for the simulations.}\label{coeftab}
   \begin{tabular}{c c c c c c}
   \hline
      & Heat  & Fisher's & AC & Sine & Tanh\\ 
     \hline 
     $\alpha$&  1&1&1&0.1&0.5\\
     $\beta$&  0&20&$6944$&40&10\\
     \hline
   \end{tabular}
 \end{center}
\end{table} 
For the AC equation, $\beta = 1/\rho^2$ where $\rho$ is the thickness of the transition layer and $\rho_5 \approx 0.012$ \cite{cnnallencahn}. For the other equations, we select arbitrary reaction coefficients. The zero Neumann boundary condition is used. For all the following equations, the continuous equations and the discretized equations are described in turn. 

{
\begin{itemize}
\item Heat equation: \begin{align}
\phi_t&=\alpha\bigtriangleup\phi. \label{heateq}
\end{align}
\item Fisher's equation: \begin{align}
\phi_t&=\alpha\bigtriangleup\phi + \beta(\phi-\phi^2). \label{fseq}
\end{align}
\item AC equation: \begin{align}
\phi_t&=\alpha\bigtriangleup\phi+ \beta(\phi-\phi^3). \label{aceq}
\end{align}
\item Reaction-diffusion equation with trigonometric function($\sin$): \begin{align}
\phi_t&=\alpha\bigtriangleup\phi + \beta \sin(\pi \phi). \label{sineeq}
\end{align}
\item Reaction-diffusion equation with trigonometric function($\tanh$): \begin{align}
\phi_t&=\alpha\bigtriangleup\phi + \beta \tanh(\phi). \label{tanheq}
\end{align}
\end{itemize}
}

When $\alpha=1$  and  $\beta=1$ , all the equations show similar evolutions, so we use different reaction coefficient $\beta$ much larger than diffusion coefficient $\alpha$ as shown in Table \ref{coeftab}.

\section{Simulation results} \label{sec:results}
\subsection{FCNNs}\label{sec:results1}
Assume that we observe a reaction-diffusion pattern and investigate the pattern rule under the constraint meaning that the observations and predictions follow the same PDE. Our proposed FCNN is trained using only two consecutive snapshots including the initial and next time step results for each equation. Then, we evaluate the model using diverse unseen initial values. 

In the simulations, we use data with random initial values with a $100 \times 100$ mesh so that the size of the input data is $102 \times 102$ containing a padding as a boundary condition. Also, $N=3$ (heat, Fisher's, AC) or $9$ (sine, tanh) for $\epsilon(\phi^n)$ is fixed depending on given equations and  a $3 \times 3$ convolutional filter is used with the stride of $1$ in Eq. \eqref{poly}. Hence, the filter has 10,000 ($=((100+2-3)/1+1)^2$) chances to learn the evolution of results images, so training a model using only two consecutive images suffices to optimize nine or thirteen model parameters ($w_0,\cdots, w_4, a_0,\cdots,a_3$).  

As an optimizer, we use ADAM \cite{adam} with a learning rate of 0.01 and without any regularization. We apply early stopping \cite{earlystopping} based on validation data to avoid overfitting. To demonstrate the approximation $\epsilon(\phi^n)$ for non-polynomial functions $f(\phi^n)$, we additionally consider sine and tanh functions in addition to heat, Fisher's, and AC equations. 

For the evaluation, we implement FCNN and FDM respectively and then measure the averaged relative $L_2$ error with 95\% confidence interval over 100 novel random initial values as shown in Table \ref{res1}. 
\begin{table}[htbp]
 \begin{center}
  \caption{Relative $L_2$ error between FCNN and FDM. The $\pm$ shows 95\% confidence intervals over 100 different random initial values.}\label{res1}
   \begin{tabular}{c c }
   \hline
     Equations & Relative $L_2$ error  \\ 
     \hline 
     Heat &  $8.4\times 10^{-5}\pm 4\times 10^{-6}$\\
     Fisher's &  $4.0\times 10^{-5}\pm 2\times 10^{-6}$ \\
     AC &  $1.3\times 10^{-6}\pm 8\times 10^{-7}$\\
     Sine &  $7.0\times 10^{-5}\pm 5\times 10^{-6}$\\
     Tanh &  $1.9\times 10^{-4}\pm 4\times 10^{-6}$\\
     \hline
   \end{tabular}
 \end{center}
\end{table} 

Furthermore, we validate the errors using different types of initial values for each equation as shown in Table \ref{simul_res2}. The initial conditions are described in the Appendix Section.
\begin{table}[htbp!]
 \begin{center}
  \caption{Relative $L_2$ error between FCNN and FDM with diverse initial values.}\label{simul_res2}
  \resizebox{\textwidth}{!}{
   \begin{tabular}{c c c c c c}
   \hline
     Initial shapes: & circle & star & circles & torus & maze\\ 
     \hline
     Heat&  $3.4\times 10^{-5}$ & $4.4\times 10^{-5}$ & $1.1\times 10^{-5}$ & $1.1\times 10^{-4}$ & $4.9\times 10^{-5}$ \\
     Fisher's&  $8.7\times 10^{-4}$ & $7.2\times 10^{-4}$ & $1.9\times 10^{-4}$ & $1.3\times 10^{-4}$ & $3.7\times 10^{-5}$\\
     AC&  $2.6\times 10^{-7}$ & $2.7\times 10^{-7}$ & $2.3\times 10^{-7}$ & $2.0\times 10^{-7}$ & $1.9\times 10^{-7}$\\
     Sine&  $3.7\times 10^{-4}$ & $2.2\times 10^{-4}$ & $9.5\times 10^{-5}$ & $7.5\times 10^{-5}$ & $4.1\times 10^{-5}$ \\
     Tanh&  $1.8\times 10^{-3}$ & $1.4\times 10^{-3}$ & $6.3\times 10^{-4}$ & $2.5\times 10^{-5}$ & $2.9\times 10^{-5}$ \\
     \hline
   \end{tabular}}
 \end{center}
\end{table}

\begin{figure}[htbp!]
\centering
\includegraphics[width=0.8\columnwidth]{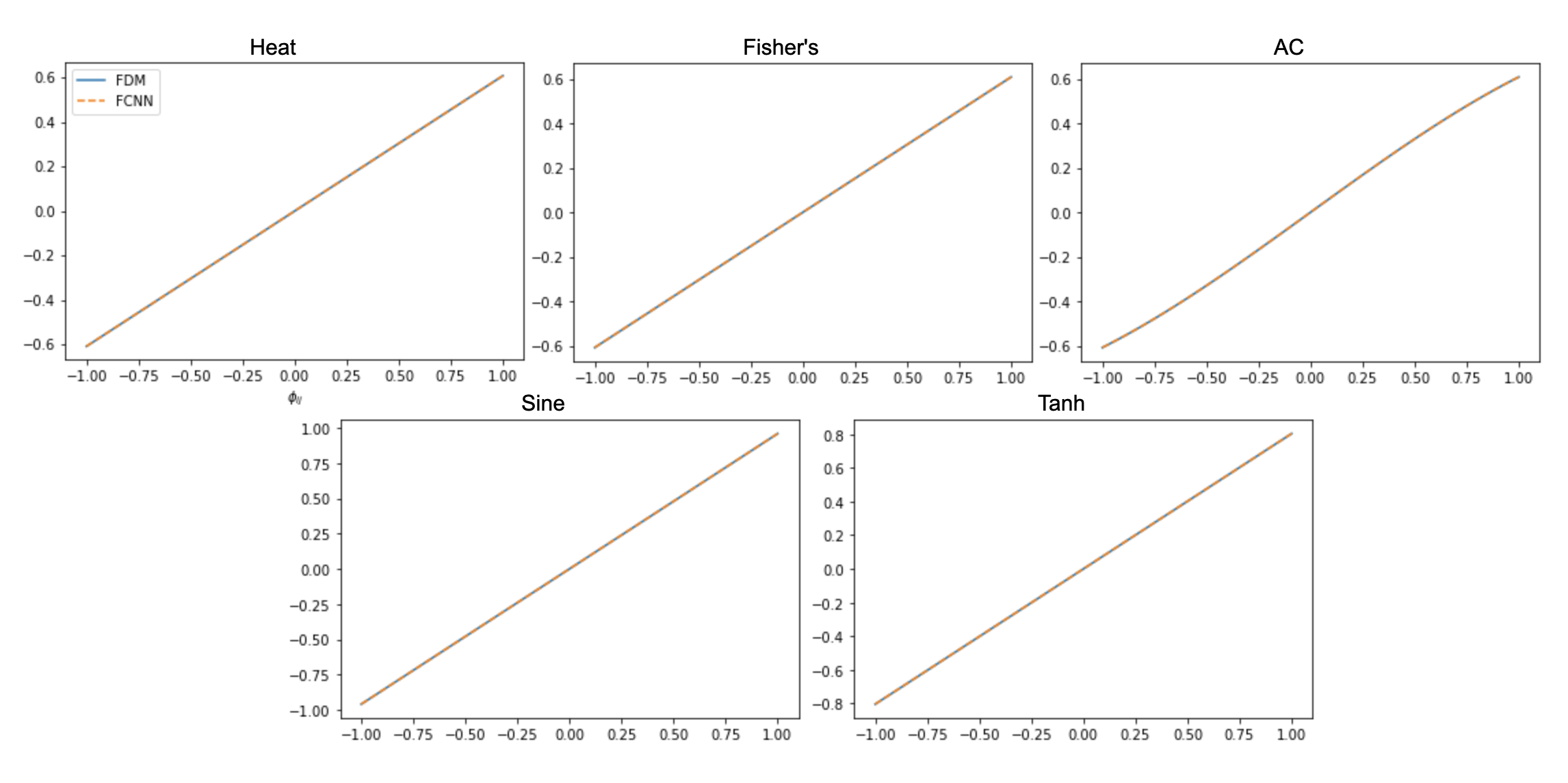} 
\caption{Center function $c(\phi_{ij})$: (blue line) FDMs, (dashed orange line) FCNNs}%
\label{coeff}
\end{figure}

As shown in Fig. \ref{coeff}, the function $c(\phi_{ij})$ calculates the $\phi_{ij}$ term in the explicit method. Also, the neighboring coefficient of the five-stencil kernel of the explicit method is ${\Delta t \alpha}/{h^2}$ and the averaged absolute error between the coefficients $w_0, w_1, w_3, w_4$ and ${\Delta t \alpha}/{h^2}$ is $1.6\times 10^{-5}$. For instance, the explicit method of the AC equation can be expressed as
\begin{equation}\label{ac_ex}
\phi_{i,j}^{n+1} = \frac{\Delta t \alpha}{h^2}(\phi_{i,j+1}^{n}+\phi_{i,j-1}^{n}+\phi_{i-1,j}^{n}+\phi_{i+1,j}^{n})+c(\phi_{i,j}^{n}),
\end{equation}
where $c(\phi_{i,j}^{n})=-4\frac{\Delta t \alpha}{h^2}\phi_{i,j}^{n}+\beta\Delta t \phi_{i,j}^{n}-\beta\Delta t (\phi_{i,j}^{n})^3$. Finally, it is shown that each numerical scheme can be reconstructed by the proposed FCNN with given $u_0$ and $u_1$.

\begin{figure}[htbp!]
\begin{minipage}{0.2\linewidth}
\raggedleft
(a)
\end{minipage}
\begin{minipage}{0.8\linewidth}
\centering
\includegraphics[width=2.7in]{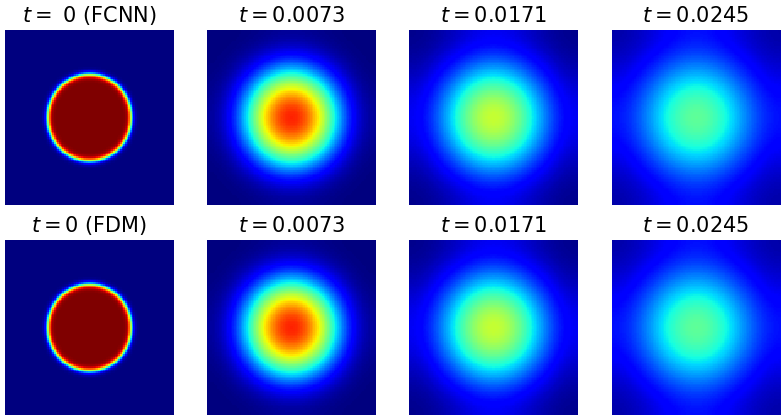}  \\
\end{minipage}\\
\begin{minipage}{0.2\linewidth}
\raggedleft
(b)
\end{minipage}
\begin{minipage}{0.8\linewidth}
\centering
\includegraphics[width=2.7in]{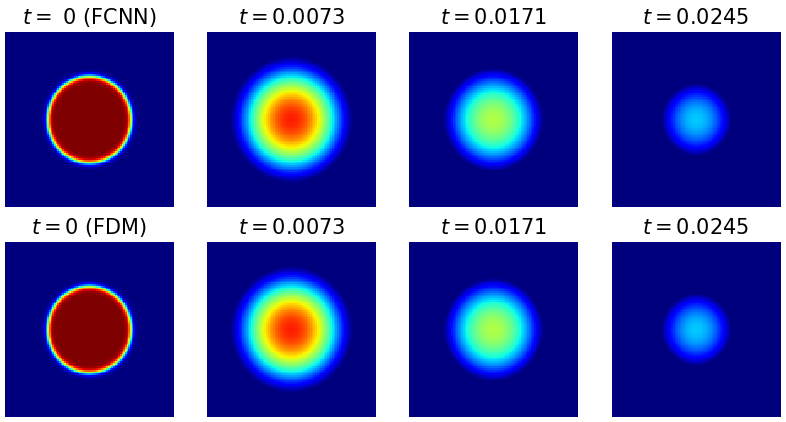} \\
\end{minipage}\\
\begin{minipage}{0.2\linewidth}
\raggedleft
(c)
\end{minipage}
\begin{minipage}{0.8\linewidth}
\centering
\includegraphics[width=2.7in]{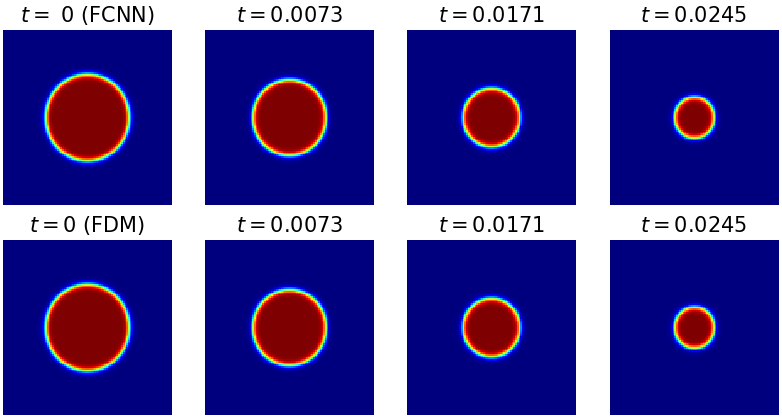} \\
\end{minipage}\\
\begin{minipage}{0.2\linewidth}
\raggedleft
(d)
\end{minipage}
\begin{minipage}{0.8\linewidth}
\centering
\includegraphics[width=2.7in]{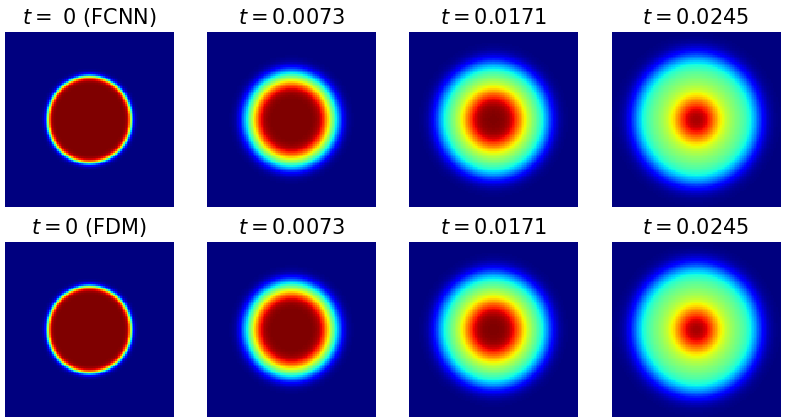} \\
\end{minipage}\\
\begin{minipage}{0.2\linewidth}
\raggedleft
(e)
\end{minipage}
\begin{minipage}{0.8\linewidth}
\centering
\includegraphics[width=2.7in]{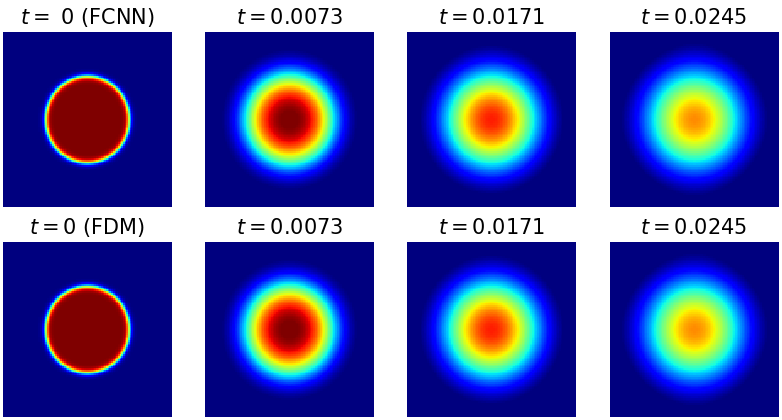} \\
\end{minipage}
\caption{Time evolution of a circle shape of (a) Heat, (b) Fisher's, (c) AC, (d) Sine, and (e) Tanh equations.}
\label{fig:image1}
\end{figure}

\begin{figure}[htbp!]
\begin{minipage}{0.2\linewidth}
\raggedleft
(a)
\end{minipage}
\begin{minipage}{0.8\linewidth}
\centering
\includegraphics[width=2.7in]{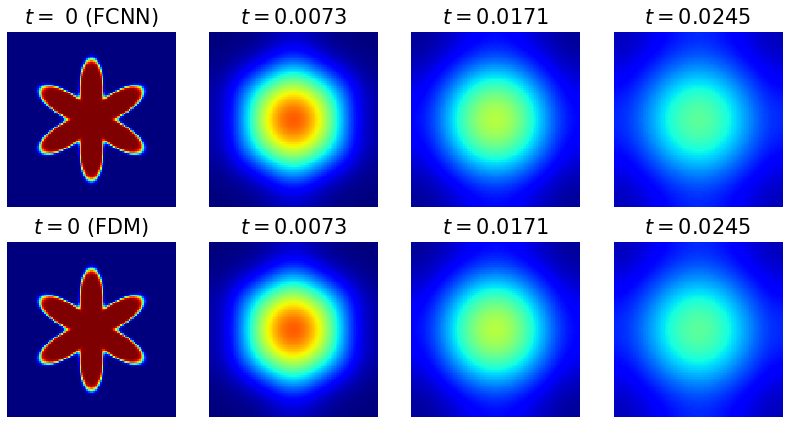}  \\
\end{minipage}\\
\begin{minipage}{0.2\linewidth}
\raggedleft
(b)
\end{minipage}
\begin{minipage}{0.8\linewidth}
\centering
\includegraphics[width=2.7in]{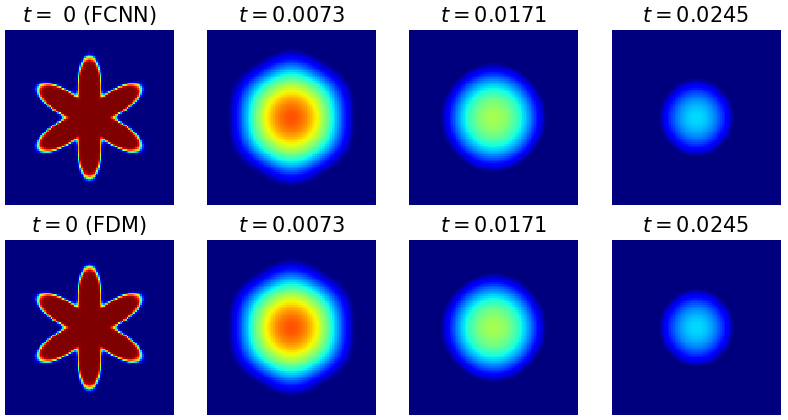} \\
\end{minipage}\\
\begin{minipage}{0.2\linewidth}
\raggedleft
(c)
\end{minipage}
\begin{minipage}{0.8\linewidth}
\centering
\includegraphics[width=2.7in]{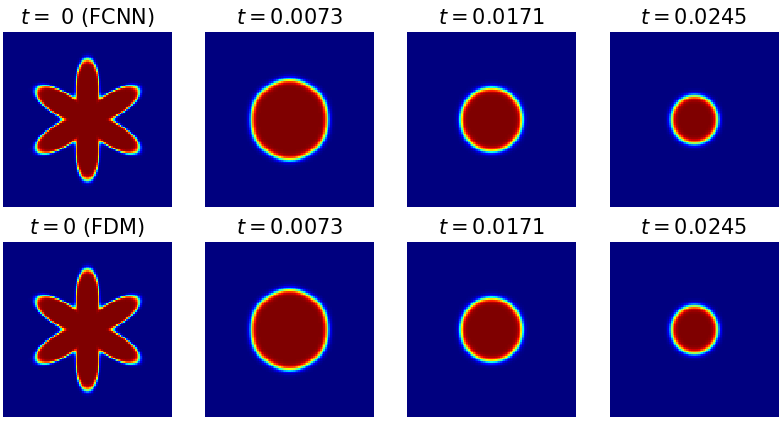} \\
\end{minipage}\\
\begin{minipage}{0.2\linewidth}
\raggedleft
(d)
\end{minipage}
\begin{minipage}{0.8\linewidth}
\centering
\includegraphics[width=2.7in]{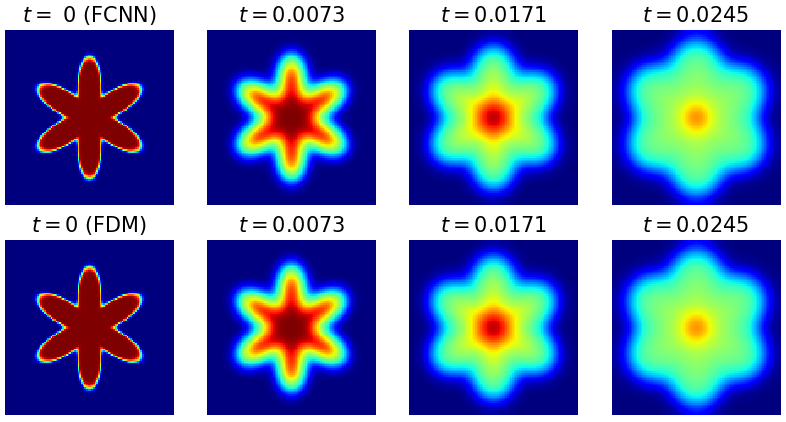} \\
\end{minipage}\\
\begin{minipage}{0.2\linewidth}
\raggedleft
(e)
\end{minipage}
\begin{minipage}{0.8\linewidth}
\centering
\includegraphics[width=2.7in]{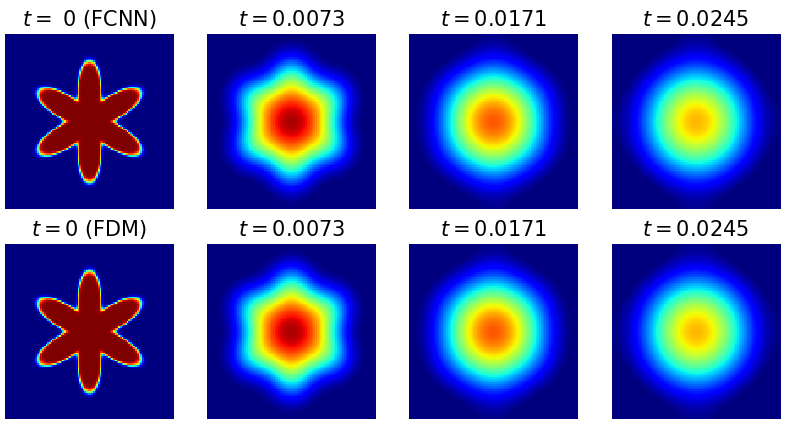} \\
\end{minipage}
\caption{Time evolution of a star shape of (a) Heat, (b) Fisher's, (c) AC, (d) Sine, and (e) Tanh equations.}
\label{fig:image2}
\end{figure}

\begin{figure}[htbp!]
\begin{minipage}{0.2\linewidth}
\raggedleft
(a)
\end{minipage}
\begin{minipage}{0.8\linewidth}
\centering
\includegraphics[width=2.7in]{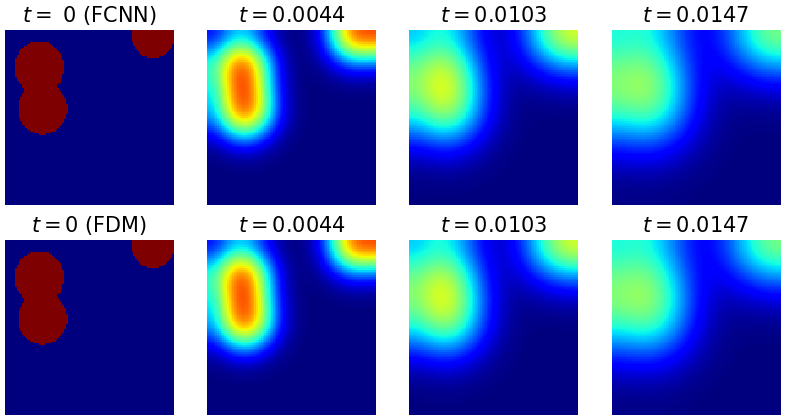}  \\
\end{minipage}\\
\begin{minipage}{0.2\linewidth}
\raggedleft
(b)
\end{minipage}
\begin{minipage}{0.8\linewidth}
\centering
\includegraphics[width=2.7in]{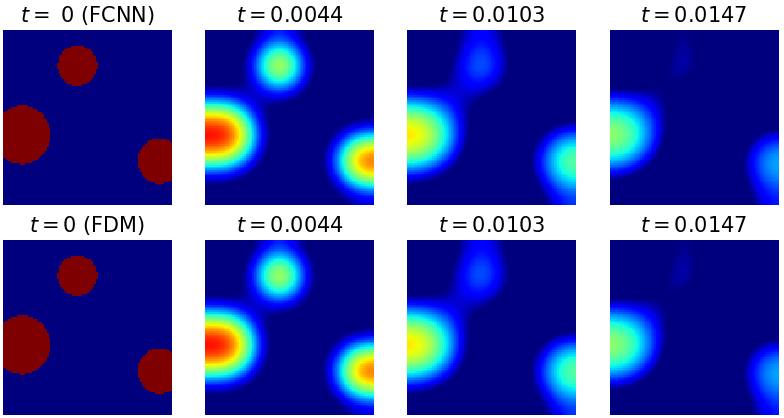} \\
\end{minipage}\\
\begin{minipage}{0.2\linewidth}
\raggedleft
(c)
\end{minipage}
\begin{minipage}{0.8\linewidth}
\centering
\includegraphics[width=2.7in]{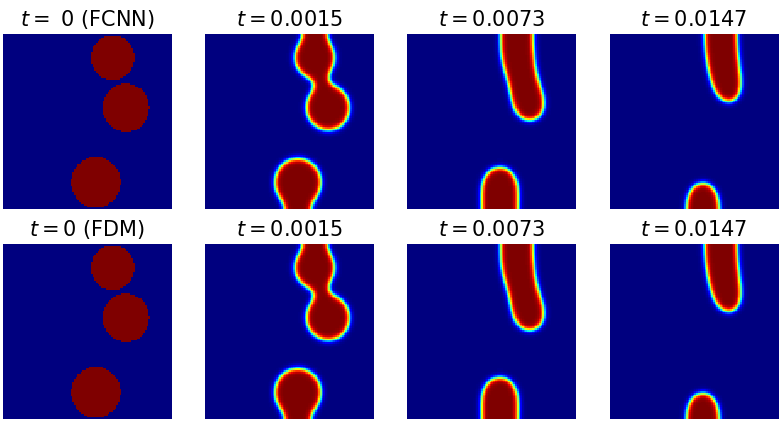} \\
\end{minipage}\\
\begin{minipage}{0.2\linewidth}
\raggedleft
(d)
\end{minipage}
\begin{minipage}{0.8\linewidth}
\centering
\includegraphics[width=2.7in]{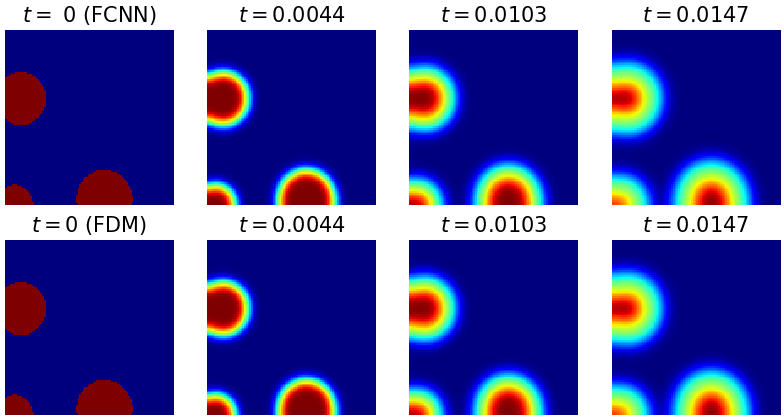} \\
\end{minipage}\\
\begin{minipage}{0.2\linewidth}
\raggedleft
(e)
\end{minipage}
\begin{minipage}{0.8\linewidth}
\centering
\includegraphics[width=2.7in]{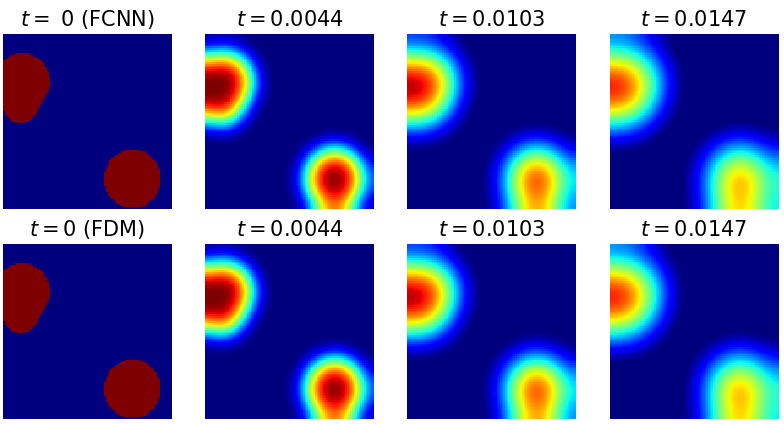} \\
\end{minipage}
\caption{Time evolution of a three circles shape of (a) Heat, (b) Fisher's, (c) AC, (d) Sine, and (e) Tanh equations.}
\label{fig:image3}
\end{figure}

\begin{figure}[htbp!]
\begin{minipage}{0.2\linewidth}
\raggedleft
(a)
\end{minipage}
\begin{minipage}{0.8\linewidth}
\centering
\includegraphics[width=2.7in]{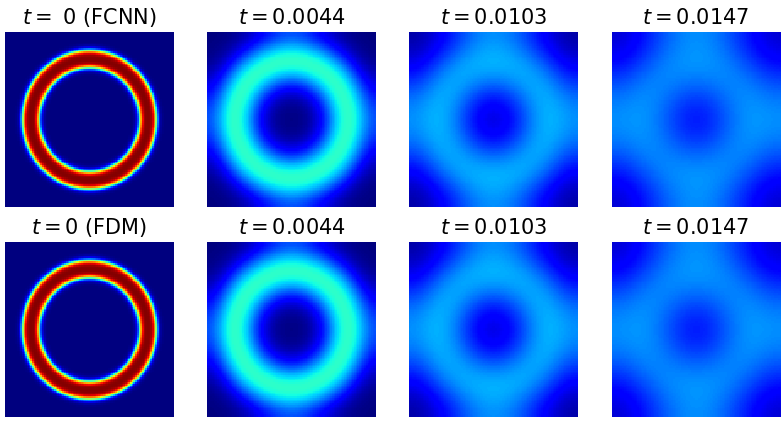}  \\
\end{minipage}\\
\begin{minipage}{0.2\linewidth}
\raggedleft
(b)
\end{minipage}
\begin{minipage}{0.8\linewidth}
\centering
\includegraphics[width=2.7in]{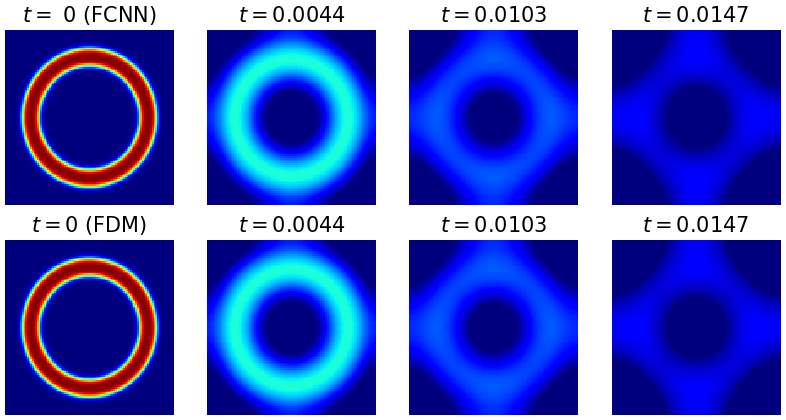} \\
\end{minipage}\\
\begin{minipage}{0.2\linewidth}
\raggedleft
(c)
\end{minipage}
\begin{minipage}{0.8\linewidth}
\centering
\includegraphics[width=2.7in]{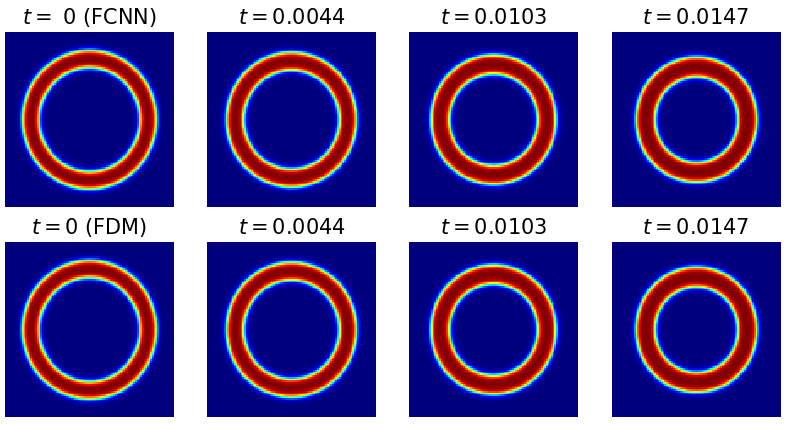} \\
\end{minipage}\\
\begin{minipage}{0.2\linewidth}
\raggedleft
(d)
\end{minipage}
\begin{minipage}{0.8\linewidth}
\centering
\includegraphics[width=2.7in]{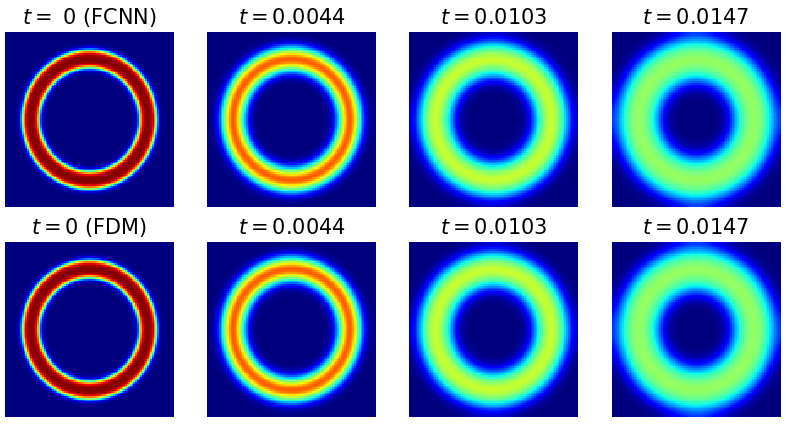} \\
\end{minipage}\\
\begin{minipage}{0.2\linewidth}
\raggedleft
(e)
\end{minipage}
\begin{minipage}{0.8\linewidth}
\centering
\includegraphics[width=2.7in]{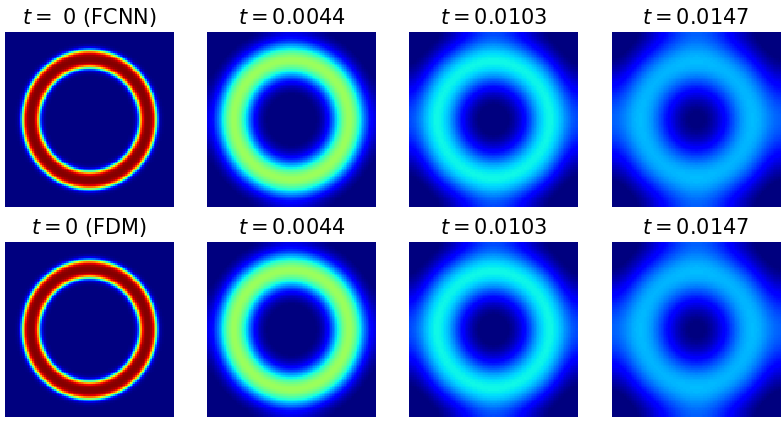} \\
\end{minipage}
\caption{Time evolution of a torus shape of (a) Heat, (b) Fisher's, (c) AC, (d) Sine, and (e) Tanh equations.}
\label{fig:image4}
\end{figure}

\begin{figure}[htbp!]
\begin{minipage}{0.2\linewidth}
\raggedleft
(a)
\end{minipage}
\begin{minipage}{0.8\linewidth}
\centering
\includegraphics[width=2.7in]{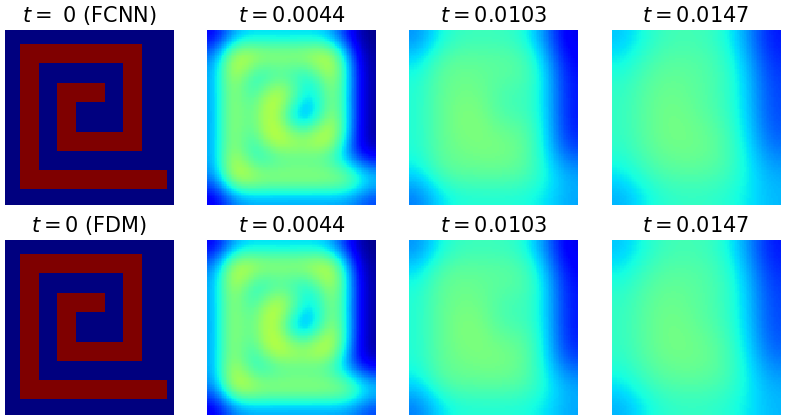}  \\
\end{minipage}\\
\begin{minipage}{0.2\linewidth}
\raggedleft
(b)
\end{minipage}
\begin{minipage}{0.8\linewidth}
\centering
\includegraphics[width=2.7in]{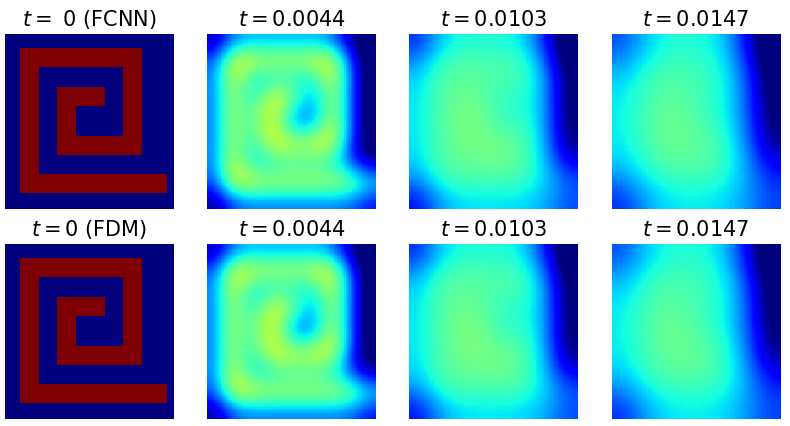} \\
\end{minipage}\\
\begin{minipage}{0.2\linewidth}
\raggedleft
(c)
\end{minipage}
\begin{minipage}{0.8\linewidth}
\centering
\includegraphics[width=2.7in]{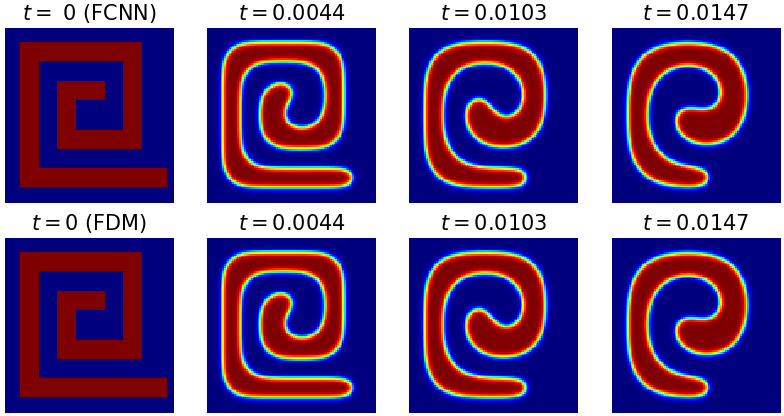} \\
\end{minipage}\\
\begin{minipage}{0.2\linewidth}
\raggedleft
(d)
\end{minipage}
\begin{minipage}{0.8\linewidth}
\centering
\includegraphics[width=2.7in]{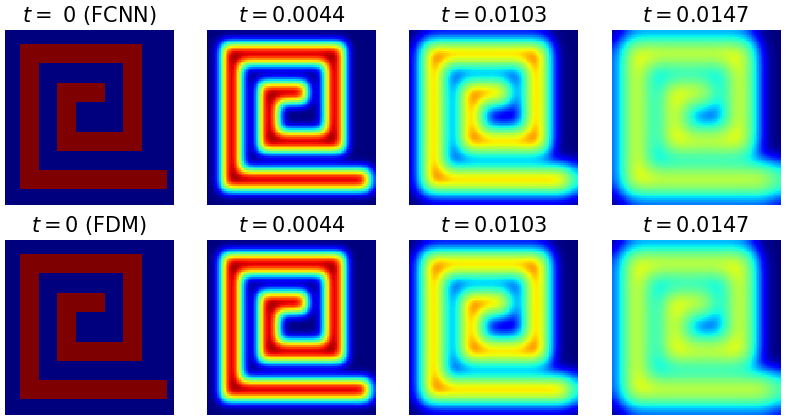} \\
\end{minipage}\\
\begin{minipage}{0.2\linewidth}
\raggedleft
(e)
\end{minipage}
\begin{minipage}{0.8\linewidth}
\centering
\includegraphics[width=2.7in]{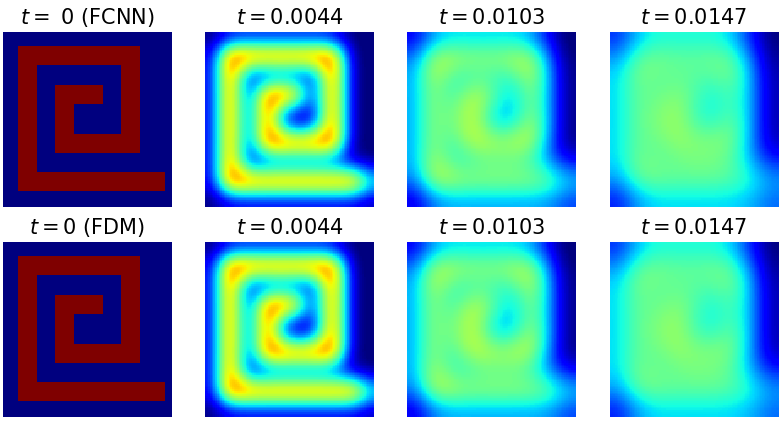} \\
\end{minipage}
\caption{Time evolution of a maze shape of (a) Heat, (b) Fisher's, (c) AC, (d) Sine, and (e) Tanh equations.}
\label{fig:image5}
\end{figure}

Figures \ref{fig:image1}-\ref{fig:image5} show the time evolution results from unseen initial shapes (circle, star, three circles, torus, and maze) using the pre-trained FCNN of each equation to compare them to the FDM results.

\newpage
\subsection{FCNNs with noisy data}\label{sec:results2}

Data-driven models are sensitive to data noise. To investigate the effects of noise on the our proposed model, we inject Gaussian random noise $\eta$ $\sim$ $N(0, \sigma^2)$ to $u^1$ and then the model is trained using $u_0$ and $u_1+\eta$ for the AC equation. Table \ref{res_noise} shows that the model could be trained under the noise condition. 
\begin{table}[htbp!]
 \begin{center}
  \caption{Relative $L_2$ error with noise. The $\pm$ shows 95\% confidence intervals over 100 different random initial values.}\label{res_noise}
   \begin{tabular}{c c | c c}
   \hline
     $\sigma$ & Relative $L_2$ Error & $\sigma$ & Relative $L_2$ Error  \\ 
     \hline 
     $0$ & $1.3\times 10^{-6}\pm 8\times 10^{-7}$ & $10^{-4}$ &  $3.3\times 10^{-4}\pm 2\times 10^{-4}$ \\
     $10^{-6}$ &  $9.1\times 10^{-5}\pm 6\times 10^{-4}$ & $10^{-2}$ &  $1.4\times 10^{-1}\pm 3\times 10^{-2}$ \\
     \hline
   \end{tabular}
 \end{center}
\end{table}

\begin{figure}[htbp!]
\centering
\includegraphics[width=0.65\columnwidth]{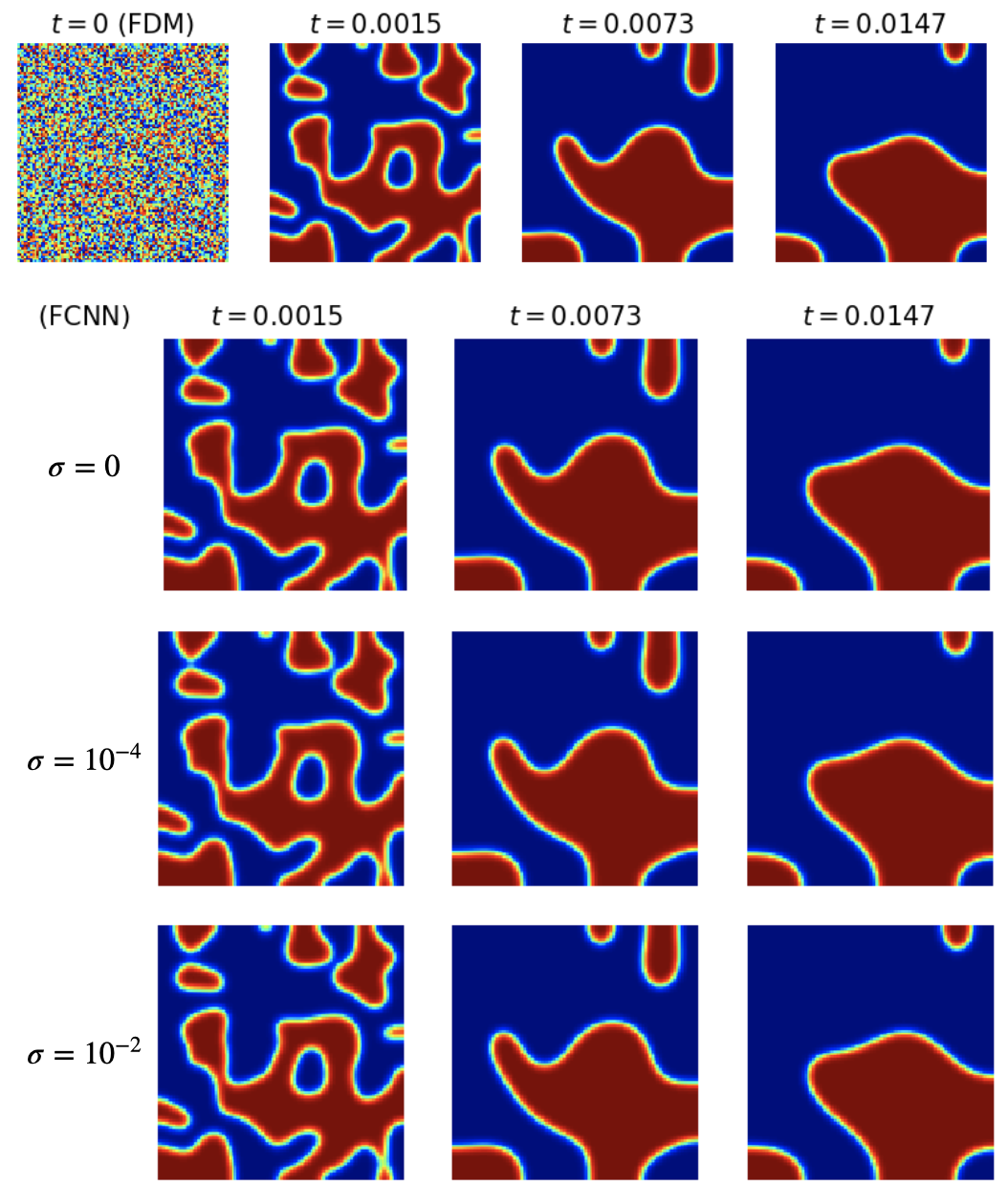}
\caption{Inference using a contaminated model with the noise impact $\sigma$}%
\label{noisemodel}
\end{figure}

\noindent Figure \ref{noisemodel} shows the results of the inference using contaminated models.

\subsection{Comparison between PINNs and FCNNs}\label{sec:results3}
\begin{table}[htbp]
 \begin{center}
  \caption{PINNs vs. FCNNs in second order reaction-diffusion type equations}\label{pinn_fcnn}
  \
   \begin{tabular}{c c c c c}
   \hline
   &  \multicolumn{2}{c}{PINNs}   &  \multicolumn{2}{c}{FCNNs} \\
     \hline
    model type&  \multicolumn{2}{c}{continuous}   &  \multicolumn{2}{c}{discretized} \\
    domain&  \multicolumn{2}{c}{mesh-free}   &  \multicolumn{2}{c}{mesh-dependency} \\
    observation&  \multicolumn{2}{c}{not required}   &  \multicolumn{2}{c}{only $u_1$} \\
    optimization&  \multicolumn{2}{c}{hard}   &  \multicolumn{2}{c}{easy} \\   
    training&  \multicolumn{2}{c}{$u_0$-dependency}   &  \multicolumn{2}{c}{$u_0$-free} \\
   \hline
   \end{tabular}
 \end{center}
\end{table}

As shown in Table \ref{pinn_fcnn}, a mesh-free and continuous-time PINN has no constraint on domain structures. Hence, it has the potential to solve PDE solutions without any discretization. Also, observation data are not required to train a model. Nevertheless, several problems remain, such as the untractable PINN optimization and the long training runtime as well as PINNs with each different initial condition $u_0$ should be trained respectively, despite considering the same PDE. On the contrary, FCNNs can solve the PDE solutions for any initial conditions using a pretrained model that learned evolution patterns from two consecutive snapshots.

We consider the heat equation (\ref{heateq}) to compare between the pretrained FCNN and the standard PINN. For the implementation of a PINN, we use 50,000 ($20\times50\times50$) collocation data, the circle initial condition $u_0$ (\ref{2Dcircleini}), with a zero Neumann boundary condition. The baseline network is a MLP consisting of an input layer, three hidden layers, and an output layer with an activation function $tanh$ in each hidden layer. Each hidden layer has 50 nodes. In the training session, we use the ADAM with a learning rate of $10^{-4}$. The loss function is defined as
\begin{align}
L=&L_c+L_b+\lambda L_{ini} \label{loss}
\end{align}
where $L_c=\sum_{i=1}^{N_c}f(t_i,x_i,y_i)^2$ is the physics-informed loss, $L_b=\sum_{i=1}^{N_b}\hat{u}(t_i,x^{b}_i,y^{b}_i)^2$ is the boundary condition loss with the coordinates $(t,x^{b},y^{b})$ on the boundary, $L_{ini}=\sum_{x,y\in\Omega_h}(\hat{u}(0,x,y)-u_0(x,y))^2$ is the initial condition loss, and $\lambda$ is a positive weight of the initial condition loss. When $\lambda=1$, it is observed that the $L_c$ and the $L_b$ converge much faster than the $L_{ini}$ causing the training to be biased towards $L_c$ and $L_b$. Thus, $\lambda=50$ is selected to alleviate the optimization issue. We perform the simulation on the following specifications: Intel (R) Core (TM) i9-10900K CPU @3.70 GHz, 128 GB RAM/NVIDIA GeForce RTX 3090.
\begin{table}[htbp]
 \begin{center}
  \caption{FCNN vs. PINN: training runtime and relative $L_2$ error}\label{pf_res}
  \
   \begin{tabular}{c c c}
   \hline
    & FCNN & PINN\\
   \hline
     Relative $L_2$ error&  $6.03\times 10^{-5}$ & $1.65\times 10^{-1}$    \\
     Training runtime (hours)&  $0$ & $7$  \\
     \hline
   \end{tabular}
 \end{center}
\end{table}

\begin{figure}[htbp!]
\centering
\includegraphics[width=0.6\columnwidth]{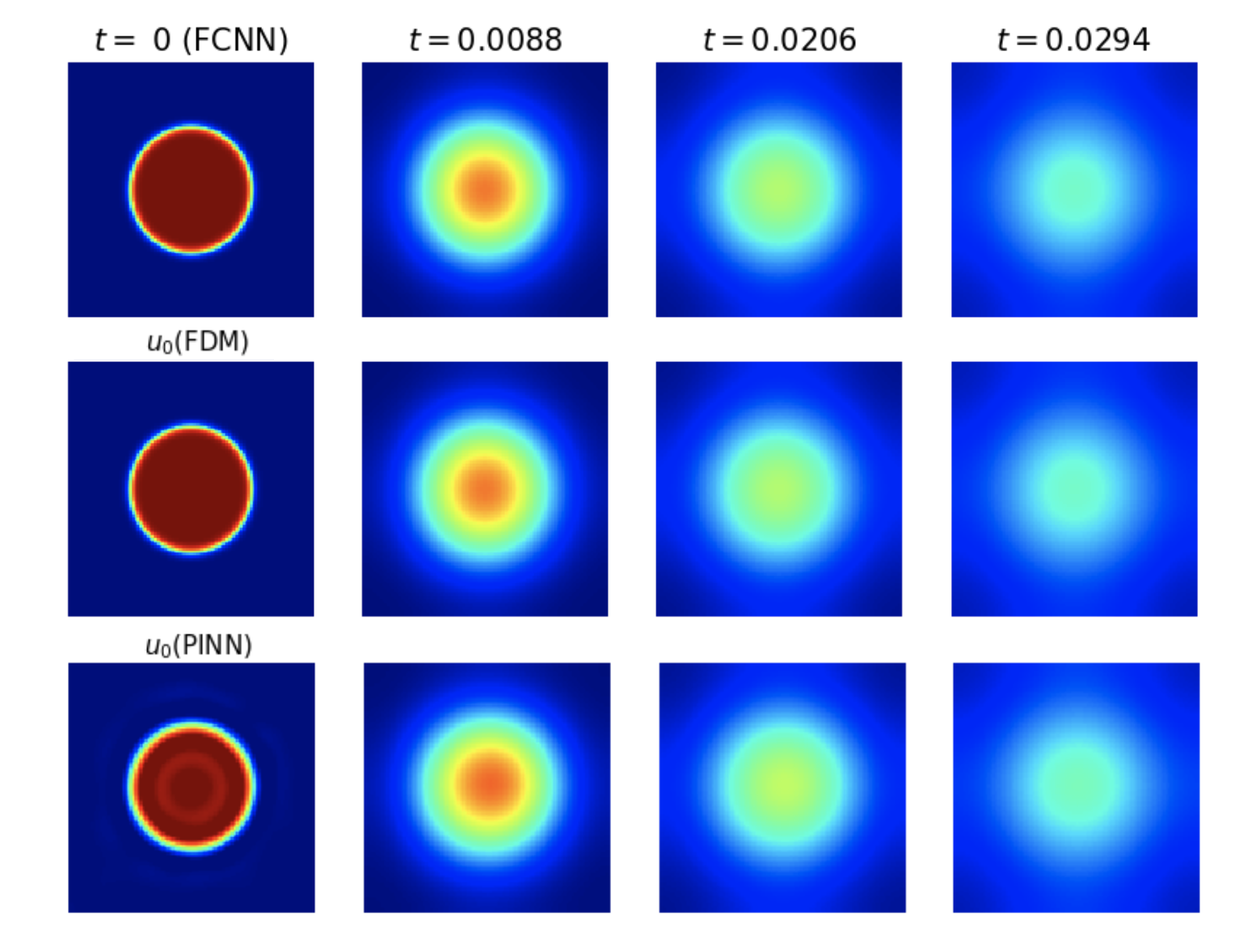}
\caption{Time evolution of the circle shape of FCNN, FDM, and PINN. PINNs even learn the initial condition from $L_{ini}$ so the trained initial values could not be exactly the same as the given values.}
\label{figpinn}
\end{figure}

\noindent Table \ref{pf_res} shows that training the PINN requires a considerable runtime, and that it is hard to optimize the model despite the expensive training cost. In contrast, the FCNN can predict the evolution without an additional training session by using the pretrained model in Section \ref{sec:results1}. It takes 2 minutes to obtain the pretrained model. The predictions of each method are shown in Fig. \ref{figpinn}.

\section{Conclusions} \label{sec:conclusions}
In this paper, we proposed Five-point stencil CNNs (FCNNs) containing a five-point stencil kernel and a trainable approximation function. We considered reaction-diffusion type equations including the heat, Fisher’s, Allen--Cahn equations, and reaction-diffusion equations with trigonometric function terms. We demonstrated that our proposed FCNN can be trained using only two consecutive snapshots and can then predict reaction-diffusion evolutions with unseen initial conditions. Also, the robustness of FCNNs was shown by the noise tests and diverse initial conditions. In future works, the characteristics of PINNs are intriguing, although the optimization of PINNs is an intractable problem. We expect that it would be feasible to train FCNNs without $u_1$ by a physics-informed loss.

\section*{Acknowledgments}
The corresponding author Y. Choi was supported by the National Research Foundation of Korea (NRF) grant funded by the Korea government (NRF-2020R1C1C1A0101153713) and supported by Basic Science Research Program through the National Research Foundation of Korea(NRF) funded by the Ministry of Education(2022R1I1A307282411). The authors appreciate the reviewers for their constructive comments, which have improved the quality of this paper.

\section*{Appendix} 
In this appendix session, we describe the initial conditions used in the simulation results session \ref{sec:results}. A detailed description of these initial conditions can be found in our previous research paper \cite{cnnallencahn}.

\noindent (1) The initial condition of a circle shape
\begin{align}
\phi(x,y,0)&=\tanh \left( \frac{R_0-\sqrt{(x-0.5)^2+(y-0.5)^2}}{\sqrt{2}\epsilon} \right),\label{2Dcircleini}
\end{align}
where $R_0$ is the initial radius of a circle.\\

\noindent (2) The initial condition of a star shape
\begin{align}
\phi(x,y,0)&=\tanh \left( \frac{0.25+0.1\cos(6\theta)-\sqrt{(x-0.5)^2+(y-0.5)^2}}{\sqrt{2}\epsilon} \right), \label{2Dstarini}
\end{align}
where 
\begin{equation*}
\theta=
\begin{cases}
\tan^{-1} \left( \frac{y-0.5}{x-0.5} \right), \quad \textrm{if}~~(x>0.5) \\
\pi + \tan^{-1} \left( \frac{y-0.5}{x-0.5} \right), \quad \textrm{otherwise}.
\end{cases}
\end{equation*}\\

\noindent (3) The initial condition of a torus shape
\begin{align}
\phi(x,y,0)&=-1+\tanh \left( \frac{R_1-\sqrt{XY}}{\sqrt{2}\epsilon} \right) - \tanh \left( \frac{R_2-\sqrt{XY}}{\sqrt{2}\epsilon} \right), \label{2Dtorusini}
\end{align}
where $R_1$ and $R_2$ are the radius of major (outside) and minor (inside) circles, respectively. And, for simplicity of expression, $XY=(x-0.5)^2+(y-0.5)^2$.\\

\noindent (4) The initial condition of a maze shape

The initial condition of a maze shape is complicated to describe its equation, so refer to the codes which are available from the first author's GitHub web page (https://github.com/kimy-de/fcnn) and the corresponding author's web page (https://sites.google.com/view/yh-choi/code).\\

\noindent (5) The initial condition of a random shape
\begin{align}
\phi(x,y,0)&=0.1\textrm{rand}(x,y), \label{2Dtrnadini}
\end{align}
here the function rand$(x,y)$ has a random value between $-1$ and $1$.

\end{document}